\documentclass[conference,letterpaper]{IEEEtran}
\IEEEoverridecommandlockouts
\usepackage{comment}
\usepackage{cite}
\usepackage{amsmath,amssymb,amsfonts}
\usepackage{algorithmic}
\usepackage{graphicx}
\usepackage{textcomp}
\usepackage[dvipsnames]{xcolor}
\usepackage{array}
\usepackage{multirow}
\usepackage{subcaption}
\def\BibTeX{{\rm B\kern-.05em{\sc i\kern-.025em b}\kern-.08em
    T\kern-.1667em\lower.7ex\hbox{E}\kern-.125emX}}

\begin{document}
	
	\title{A Deep Neural Network for Finger Counting and Numerosity Estimation}
	
	\author{\IEEEauthorblockN{Leszek Pecyna}
	\IEEEauthorblockA{\textit{Centre for Robotics and Neural Systems} \\
	\textit{University of Plymouth}\\
	Plymouth, United Kingdom \\
	leszek.pecyna@plymouth.ac.uk}
	\and
	\IEEEauthorblockN{Angelo Cangelosi}
	\IEEEauthorblockA{
	\textit{University of Manchester and}\\
	\textit{ICAR-CNR Institute}\\
	Manchester, United Kingdom \\
	angelo.cangelosi@manchester.ac.uk}
	\and
	\IEEEauthorblockN{Alessandro Di Nuovo}
	\IEEEauthorblockA{\textit{Sheffield Robotics} \\
	\textit{Sheffield Hallam University}\\
	Sheffield, United Kingdom \\
	a.dinuovo@shu.ac.uk}
	}
	
	\maketitle
	
\begin{abstract}
    In this paper, we present neuro-robotics models with a deep artificial neural network capable of generating finger counting positions and number estimation. We first train the model in an unsupervised manner where each layer is treated as a Restricted Boltzmann Machine or an autoencoder. Such a model is further trained in a supervised way. This type of pre-training is tested on our baseline model and two methods of pre-training are compared. The network is extended to produce finger counting positions. The performance in number estimation of such an extended model is evaluated. We test the hypothesis if the subitizing process can be obtained by one single model used also for estimation of higher numerosities.
    The results confirm the importance of unsupervised training in our enumeration task and show some similarities to human behaviour in the case of subitizing.
    
    \end{abstract}
    
    
    \section{Introduction}

    There is strong evidence that gestures like pointing or finger counting have a positive contribution to numerical cognition development in children\cite{goldin2014gestures,alibali1999function,fischer2012finger,andres2008finger}.
    Finger counting has been hypothesised to help in acquiring mechanisms such as one-to-one correspondence, stable order, and cardinality which are fundamental to the development of counting system \cite{domahs2012handy}. It has been observed that the use of finger counting to support mathematical learning and to achieve better performance changes over time. It is not used by children after they learned the basic concepts \cite{jordan2008development}.
    
    Another important aspect of numerosity estimation is subitizing. This is an effortless, fast and accurate enumeration process, concerning a small number of objects up to around 4. Beyond this range, the numerosity naming becomes much slower with a linear increase of 200 to 400 ms/item \cite{trick1994small}. It has been demonstrated that participants can approximately estimate the number of objects without counting and this estimation follows Weber's law (the answers are increasingly less precise and the variability increases proportionally to the mean response)\cite{izard2008calibrating,revkin2008does}.
    As this rule does not apply for small numerosities (1-4) there is a hypothesis that there are two dedicated numerical estimation systems for small and large sets in contrast to the hypothesis that there might be one single system for both sets \cite{revkin2008does}.
    
    
    A single system made from a generic neural network for numerosity estimation where the mentioned hypotheses have been challenged has been shown in \cite{chencan}. The performance of the model was analysed depending on the numerosity of presented objects.
    This provided some insight into the ability of the network (trained in presented there supervised way) for both subitizing and numerosity estimation (of higher numbers following Weber's law).
    
    An unsupervised training, where the network is trained as a Restricted Boltzmann Machine (RBM) \cite{hinton2006fast}, has been shown to be successful for numerosity visual sensation and numerosity comparison tasks \cite{stoianov2012emergence,zorzi2018emergentist}. It was also shown that it boosts the training when used as pre-training in a Deep Neural Network (DNN) for tasks related with number cognition \cite{DiNuovo2015}. Autoencoders used in a similar way have been shown to improve the training performance as well \cite{di2017embodied}. In both of those publications the networks were performing number cognition related task 
    (but not a numerosity estimation task as in the present work).
    
    As mentioned before, there is evidence for the importance of gestures in numerical cognition development in children. Thus, Cognitive Developmental Robotics (CDR) is naturally suitable to study embodied basis of mathematical learning \cite{di2017embodied,di2019development}, where the robot is used to interact with the environment and where sensorimotor data can be used in the learning process. Robotic platforms have been used for finger counting in several models \cite{Vivian2014,DiNuovo2014,DiNuovo2014a,DiNuovo2015,di2017embodied,shu20884}. 
    While in some others it was also used for pointing \cite{rucinski2012robotic,pecyna2018influence}.

    The purpose of our experiments is to check if a relatively simple neural network can show some similarities in numerosity estimation development with children when trained in a particular way. We investigate the importance of unsupervised pre-training and the influence of motor input.
    In this paper, we present a variety of training simulations and analyses using robotic embodiment,  deep neural network, unsupervised and supervised training for numerosity estimation task.
    The embodiment for our experiments is provided by an iCub humanoid robot.
    The data from iCub hands is used as a sensorimotor signal for finger counting. 
    In particular, we provide some comparison of performances of two types of unsupervised pre-training (RBM and autoencoder) used prior to the main one. We train the model to produce the gestures itself and we check how this influences the enumeration learning process.
    As in \cite{chencan}, we analyse the output from our network to check if one system made from a generic neural network could be capable of subitizing and numerosity estimation (but using embodiment, and unsupervised pre-training). Thus, it is interesting to observe how those training (and architecture) differences influence the results. As in \cite{chencan}, we use two types of number distributions during the training (uniform and based on a more realistic appearance of numbers in a human environment).
    
    The paper is organised as follows. First, we present the baseline model where both motor and visual inputs are used. The model allows us to chose proper architecture for further experiments and compare different pre-training approaches. Next, we extend the model so that the network is producing the sensorimotor data itself. On this model, we perform some experiments to show if the training with this additional motor output performs better compared to the training with only enumeration output. In the following section, we are analysing the data with regard to the development of numerosity estimation for a different number of objects paying additional attention to the subitizing process. The article finishes with our conclusions and future work discussion.

    \section{Baseline model with motor and visual inputs}\label{baseline}
    \begin{figure}[tb]
        \centerline{\includegraphics[width = 0.8\columnwidth]{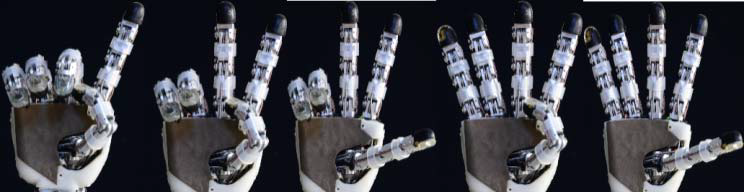}}
        \caption{Number representation using iCub right hand fingers. The joints' angles are used as a motor input (figure from \cite{DiNuovo2014a}).}
        \label{fig:Hand}
    \end{figure}
    The model training is based on \cite{DiNuovo2015} with some changes and simplifications added to the architecture and described in Section \ref{model structure}. Another difference is that our model is used for the visual enumeration task (the model in \cite{DiNuovo2015} was used for recognition of number words in auditory input). In the works of Di Nuovo et al. \cite{DiNuovo2014,DiNuovo2014a,DiNuovo2015,di2017embodied,shu20884} the number cognition tasks (recognition of number names from auditory input or from written numbers) are supported by a motor module that provides the fingers' motor representation of the numbers from robotic or robot simulator hands (see Fig.~\ref{fig:Hand}). In our model, we use a similar support module.
    
    The network is first pre-trained in a similar way as in \cite{DiNuovo2015, di2017embodied}. In the model presented in \cite{di2017embodied}, the link was added between the visual module and the motor module. In the following sections, a similar link where the motor input is produced by the network is also implemented.

        \subsection{Model description}
        The model is trained to produce the name corresponding to the numerosity based on the provided visual and/or motor input. It is supposed to answer the question of how many objects are present in the visual input or how many fingers are straightened.
        
        As mentioned before we use an iCub humanoid robot as an embodiment providing us sensorimotor data. It is a childlike robot, 1.05 m tall with 53 degrees of freedom (DoF). More about the robot can be found in \cite{metta2010icub}. The iCub platform was used because of its complex hands' architecture (9 DoF each) that allows producing human-like gestures.

        \subsubsection{Model architecture}\label{model structure}

        \begin{figure}[tb]
            \centerline{\includegraphics[width = 0.85\columnwidth]{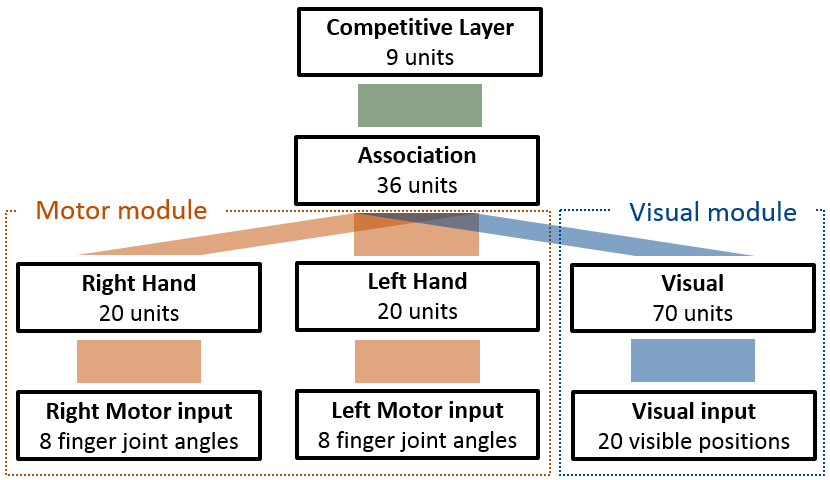}}
            \caption{Architecture - grey polygons represent all-to-all connections between the layers of neurons.}
            \label{fig:Architecture_final}
        \end{figure}
        The architecture of the tested model can be found in Fig.~\ref{fig:Architecture_final}. As in \cite{DiNuovo2015}, we used motor module with the Right Hand and Left Hand layers (to resemble the activity of the human brain - two hemispheres). In our case, this module is one layer shallower and the auditory input was replaced by the visual input.
        The architecture of our model is composed of two main modules, the Association layer (which combines motor and visual modules, similarly to ``Motor-Auditory association'' layer in \cite{DiNuovo2015}) and the Competitive layer (using Softmax activation function) which is producing the output corresponding to the numerosity estimation.

        \subsubsection{Inputs and outputs}\label{input}
        There are three inputs to the network (two in the motor module and one in the visual module) and one output. There are described as follows:
        \begin{itemize}
            \item Right Motor input: It consists of 8 units representing collected angles of joints in the iCub robot right hand. The values of the input for each number are the same as in Table 3 and 4 in \cite{DiNuovo2014a} for numbers from 1 to 9. Because ring and pinky fingers are controlled together (are "glued" together) the fingers have 7 DoF for each hand. To balance the input from motors controlled together, we duplicated their angles units (ring and pinky fingers) in the input. The same approach was used in \cite{di2017embodied}.
            \item Left Motor input: Corresponding input as the one described in the previous point but for the left hand.
            \item Visual input: This is a 1-dimensional vector (saliency map), which can be considered a simple model of the retina, it consists of 20 units corresponding to 20 spacial locations. The visual input is the same as described in \cite{pecyna2018influence,rucinski2012robotic}. Depending on the presence of an
            object in a given location, each unit will be activated or not. The maximal number of objects assumed in the presented experiments is 9, the locations of objects are randomized. The visual input vector is normalized in order to eliminate the possibility of discrimination of cardinality based on summed activation of that input.
            \item Competitive Layer: This is a 1-dimensional vector with 9 units where one-hot coding has been used to represent each of the numbers.
        \end{itemize}
        
        \subsection{Training description and variations}

        The neural network is pre-trained layer-by-layer in an unsupervised way using the same method as in \cite{DiNuovo2015, di2017embodied} (the description of that method can be found in \cite{hinton2006fast}). In \cite{DiNuovo2015} and \cite{di2017embodied} articles, the models are pre-trained using RBM or autoencoder training techniques respectively, our model will be trained in both ways so that we can see if both techniques provide additional improvement in the training (and how different they are from one another).
        The complex comparison between RBMs and autoencoders can be found in \cite{testolin2017role}, however, for a different task (making a compact representation of eye and effector positions) and using different architecture.
        
        The model was tested in 3 architecture configurations:
        \begin{itemize}
            \item Only visual module used
            \item Only motor module used
            \item Both modules used
        \end{itemize}
    
        Each configuration was tested in 3 training modalities:
        \begin{itemize}
            \item Pre-trained as RBMs
            \item Pre-trained as autoencoders
            \item No pre-training
        \end{itemize}
        
        We trained the model for numbers in the range from 1 to 9.
        All training data was composed in mini-batches. Each mini-batch was a collection of numbers (from 1 to 9). After some trials, the visual training data was chosen to contain 500 randomly generated mini-batches (making one epoch). Pre-training was run over 1500  iterations (3 epochs) in all simulations, this has been found to be enough and further extension of it was not providing significant improvement. The test set was composed of 500 batches.
        
        We used an Adam optimizer \cite{kingma2014adam} for both the main training and the pre-training with autoencoders. For RBM training we based our approach on the network described in \cite{Zorzi2005,testolin2017role}. We implemented the model in Python using Tensorflow.
        
        The parameters were chosen using hyperparameter search in the following way. In the first stage, we used only the visual module and the parameters for this part of the network were chosen. This includes sizes of the hidden layers and learning rate for main training and pre-training. Next, we run the tests using the motor module and chose the parameters for that module. Finally, we adapted the learning rates values for the network with both modules. The sizes of the hidden layers are presented in Fig.~\ref{fig:Architecture_final} and the values of selected learning rates can be found in Table~\ref{learning rate}.
        \begin{table}[tb]
        \caption{Values of learning rates for the training of our model}
        \begin{tabular}{>{\centering}m{0.09\columnwidth}
                >{\centering}m{0.16\columnwidth}
                >{\centering}m{0.17\columnwidth}
                >{\centering}m{0.17\columnwidth}
                >{\centering\arraybackslash}m{0.17\columnwidth}}
            \hline
            \textbf{Modules} && \textbf{RBMs} & \textbf{Autoencoders} & \textbf{No pre-training} \\
            \hline
            \multirow{2}{*}{\shortstack{Visual}}&Pre-training& 0.3 & 0.02 & - \\ 
            \cline{2-5}
            &Final& 0.004 & 0.004 & 0.01 \\ 
            \hline
            \multirow{2}{*}{\shortstack{Motor}}&Pre-training& 0.4 & 0.08 & - \\ 
            \cline{2-5}
            &Final& 0.2 & 0.13 & 0.03 \\ 
            \hline
            \multirow{2}{*}{\shortstack{Both}}&Pre-training& 0.4 & 0.04 & - \\ 
            \cline{2-5}
            &Final& 0.1 & 0.11 & 0.02 \\ 
            \hline     
        \end{tabular}\label{learning rate}
        \end{table}
        
        \subsection{Results}\label{res_baseline}
            
        \begin{figure}[tb]
            \centerline{\includegraphics[width = 0.82\columnwidth]{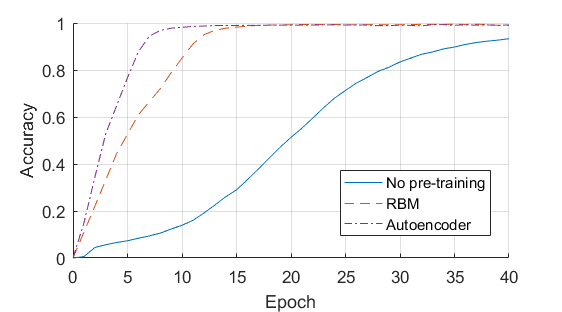}}
            \caption{The accuracy development by epoch obtained on the test set for the network using only visual module (motor module was not used, see Fig.~\ref{fig:Architecture_final}).}
            \label{fig:visual_only}
        \end{figure}
         \begin{figure}[tb]
	    	\centerline{\includegraphics[width = 0.82\columnwidth]{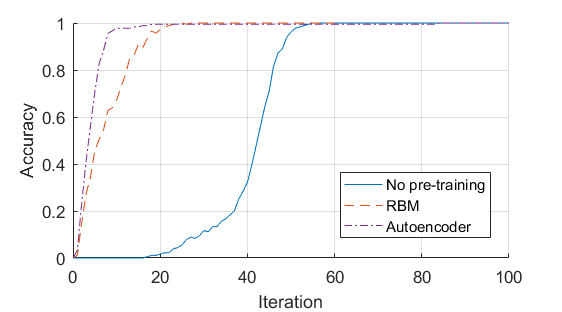}}
	    	\caption{The accuracy development by iteration obtained on the test set for the network using only motor module (visual module was not used, see Fig.~\ref{fig:Architecture_final}).}
	    	\label{fig:motor_only}
	    \end{figure}
	    
	    \begin{figure}[tb]
	    	\centerline{\includegraphics[width = 0.82\columnwidth]{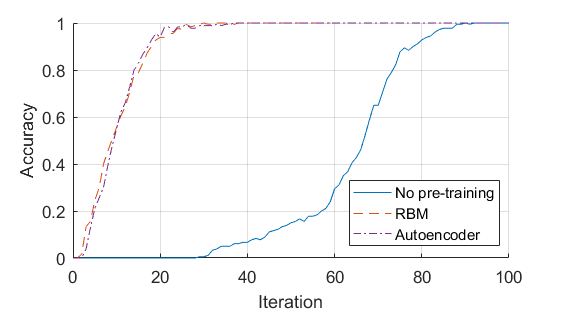}}
	    	\caption{The accuracy development by iteration obtained on the test set for the network using both modules (see Fig.~\ref{fig:Architecture_final}).}
	    	\label{fig:Full_HPS}
	    \end{figure}
        As it can be seen the best results are obtained for autoencoders pre-training and the worst when no pre-training was used.
        After each epoch (or after a specific number of iterations) the test set was generated (new random positions of objects for each numerosity in visual input) and the values in the plot where obtained. Autoencoders pre-training gave better results when training the visual module (Fig.~\ref{fig:visual_only}) or motor module only  (Fig. \ref{fig:motor_only}). In the case where both modules are used this difference is not so visible (Fig.~\ref{fig:Full_HPS}). In these charts, each curve is obtained from 20 simulations and it consists of points being average values from those runs (further plots in this article presenting accuracy development are also obtained in this way).
        The accuracy is calculated as the percentage of numbers correctly recognised, i.e. a neuron in the last layer corresponding to the correct numerosity produces the highest value.
        
        Additionally, when comparing Fig.~\ref{fig:motor_only} and Fig.~\ref{fig:Full_HPS} it can be found that the visual module, when used together with the motor module, causes a disruption for the network, decreasing its performance. This was not visible in case of RBM pre-training where the curve for only the motor module and full network were almost the same.
        
\begin{figure}[tb]
            \centerline{\includegraphics[width = 0.82\columnwidth]{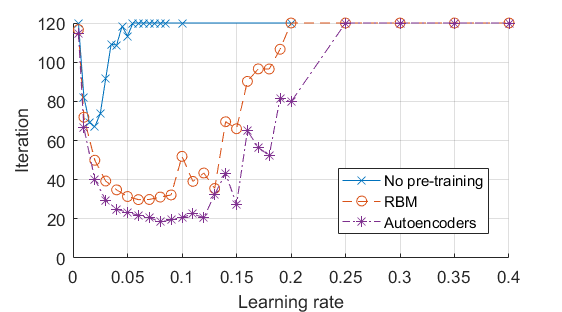}}
            \caption{Both modules used - number of iterations needed to reach accuracy of 99\% as a function of learning rate.}
            \label{fig:visual_only_time}
        \end{figure}

        Because the learning rate value is different for each training condition, epochs are not directly comparable. To overcome this problem and to make results more comparable we presented them as the number of epochs needed to converge to 99\% accuracy as a function of the learning rate (see Fig.~\ref{fig:visual_only_time}). In this plot, it can be found that pre-trained networks are performing much better with an autoencoder based pre-training over-performing RBM one. Each point on the figure comes from 10 simulations from which an average number of epochs needed to converge to 99\% accuracy was calculated.
        
        Generally, it can be found that training with sensorimotor input is much faster then training where only visual input was used. The first one converges to 99-100\% accuracy in less than an epoch (20-40 iterations, when pre-training used) where the second needs 10-20 epochs to converge (1 epoch is equal to 500 iterations). That comes from the fact that the visual input is randomised (random positions of objects) and the motor input gives a representation of finger positions directly corresponding to a specific number.

        Our results are in line with those presented in \cite{DiNuovo2015,di2017embodied} where also the improvement has been found when the network is pre-trained first in an unsupervised way.
        
        The results suggest better performance when using autoencoders for our task, thus, further experiments will be performed using only this type of technique.

    \section{Model that generates internal representation of finger positions} \label{gestures_out}
    The purpose of this model is to allow the network to reproduce the finger positions by itself, instead of receiving them as an external input. This approach can be considered more resembling human behaviour because children will open the fingers by themselves, and also because the representation might be disturbed and not perfect (when motor positions are used as input their representation is ideal). In this section, we are meant to show not only that the motor input increases the performance of the network but also that the network which produces finger movement itself performs better.
    
        \subsection{Model description}\label{gest_descr}
        The model was extended by adding the link between the visual module and the motor module. The network is trained to replicate the motor representation.
        
            \subsubsection{Model architecture}
\begin{figure}[tb]
                \centerline{\includegraphics[width = 0.85\columnwidth]{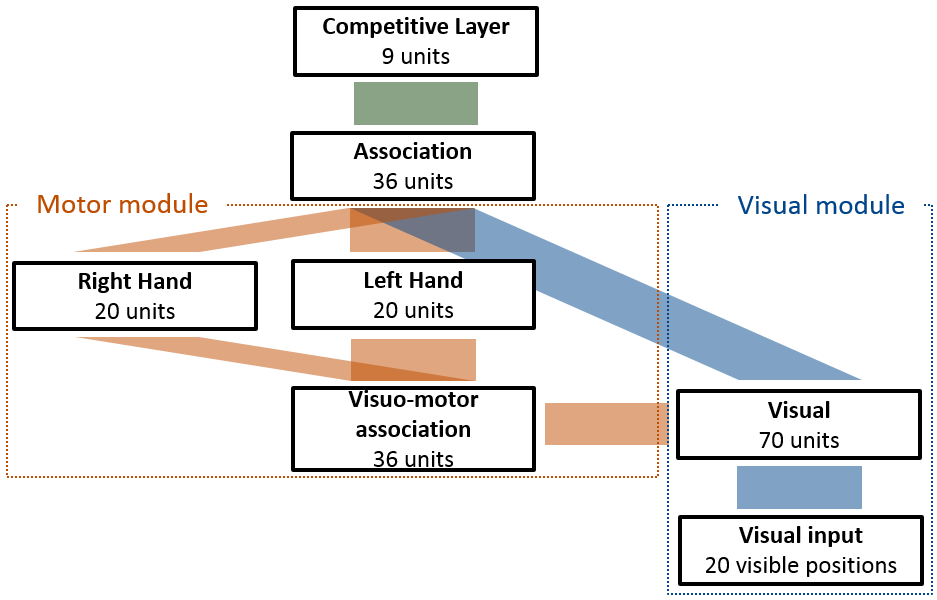}}
                \caption{Architecture - grey polygons represent all-to-all connections between the layers of neurons. Visual input is the only external input, motor input is replicated through motor module.}
                \label{fig:Full_motor_repl}
            \end{figure}

            A Visuo-motor association layer was added to the network presented in Fig.~\ref{fig:Architecture_final}. Additionally, this architecture is different as the outputs from the Right Hand and Left Hand layers are also network trainable outputs. The architecture of the network can be found in Fig.~\ref{fig:Full_motor_repl}
            \subsubsection{Inputs and outputs}\label{in_out_motor}
            There is one input to this network and one or three outputs (depending on the training). There are described as follows:
            \begin{itemize}
                \item Visual input: The same input as in the baseline model, see Section \ref{input}. It is a 1-dimensional vector with 20 units corresponding to spacial locations.
                \item Competitive Layer: The same output as in the baseline model, see Section \ref{input}. One-hot coding for 9 output numbers.
                \item Right Hand output: This is a 1-dimensional vector with 20 units corresponding to the internal representation of right hand positions. The training data for this output was obtained after the pre-training process of a baseline model. The network was fed with sensorimotor data (Right Motor input) and the Right Hand layer was trained like an autoencoder, the output data produced by the Right Hand layer is now used as the training data for the Right Hand output.
                \item Left Hand output: corresponding output as the one described in the previous point but for the left hand.
            \end{itemize}    
            The Right and Left Hand layers are not used as outputs in the preliminary stage of our experiment.
            
            \subsection{Training description}\label{train_descr2}
            For this model, we applied first the autoencoder based pre-training to train the Visual and Association layers. As shown in Section \ref{res_baseline} the autoencoder gave the best results. The Association layer was pre-trained in two ways:
            \begin{itemize}
                \item Only its visual module part was pre-trained and the rest was randomly initialized.
                \item It was pre-trained in full using inputs from the Visual layer and training data from the Right Hand and Left Hand layers (from previous architecture).
            \end{itemize}
             We used the first condition to make a baseline case (for this architecture) where the network was both not pre-trained and not trained with any sensorimotor data (but has exactly the same architecture).
            
            In the preliminary stage of our experiment, we assumed that the Right Hand, Left Hand, and Visuo-motor association layers can be pre-trained using the values of the weights from the Association layer (both from decoder and encoder) from the pre-trained network described before (Fig.~\ref{fig:Architecture_final}). Later, the model was trained only to produce enumeration output (see ``Association layer weights copied'' in Fig.~\ref{fig:Dev_dec_compare}).
            The idea under this experiment was that during the autoencoder pre-training for the Association layer the network learns to replicate the input composed of both visual and motor components, giving the network capability to replicate (at least partially) the motor input based on only visual input. We compared those results with the results from baseline architecture and with the first Association layer pre-training condition (only visual module part pre-trained).
            
            
            
            We used Adam optimizer with pre-training learning rate set to 0.04 (same as for both modules conditions in Table~\ref{learning rate}) and final learning rate set to 0.003 (adjusted using hyperparameter search).
            
            In the next stage of our experiment, the Left Hand and Right Hand layers were used as trainable outputs to check if this change can give any benefit to the training. We first trained the network to replicate the motor internal values. At this stage, the network was not trained to produce numbers. We found 6 epochs to be enough for this training with the learning rate set to 0.03. Then we performed training where the network was trained only to produce the numbers.
            
            One could conclude that in case of that training the total supervised training is longer, thus, such type of training is not necessarily beneficial. Because of that, in the last stage of our experiment, we tried to train both outputs (motors and numbers) in parallel. The network is trained both to produce the numbers at the output and to reproduce internal motor values. Tensorflow allows defining several loss functions with different learning rates that can be grouped and optimized during training. Thus, the learning rate for number training was set to 0.003 (as before); the learning rate for the motor training is a hyperbolic tangent function that starts from 0.01 and converges almost to zero in around 600 iterations. It can be described by the equation:
            \begin{equation}
            f(x)=0.01\frac{1-\tanh(\frac{6x}{600}-3)}{2}
            \end{equation}
            We chose this approach to imitate the learning process observed in children, as they use finger counting much more at the beginning of it.

        \subsection{Results} \label{gest_results}
        \begin{figure}[tb]
        	\centerline{\includegraphics[width = 0.82\columnwidth]{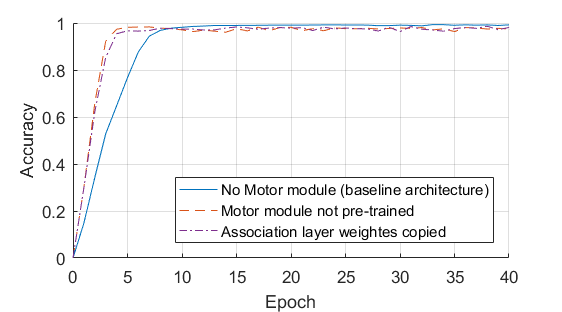}}
        	\caption{The accuracy development by epoch obtained on the test set for the network in preliminary studies. Motor output was not used in the training.}
        	\label{fig:Dev_dec_compare}
        \end{figure}
        Analysing the results of the preliminary stage of our experiment (see Fig.~\ref{fig:Dev_dec_compare}) we found that the training results are better for new more complex architecture compared to the architecture with only Visual module from baseline architecture (described in Section \ref{baseline}). However, the training that uses the Association layer weights from pre-training (as described in Section \ref{train_descr2}), did not give any improvement in the network performance, see Fig.~\ref{fig:Dev_dec_compare}.

        \begin{figure}[tb]
            \centerline{\includegraphics[width = 0.82\columnwidth]{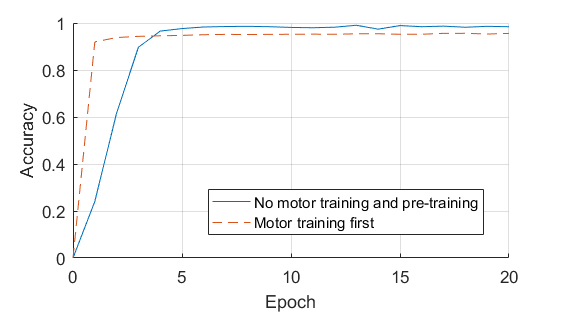}}
            \caption{The accuracy development by epoch obtained on the test set for the network that was first trained to reconstruct motor positions.}
            \label{fig:Reconstr_first}
        \end{figure}
	    \begin{figure}[tb]
	    	\centerline{\includegraphics[width = 0.82\columnwidth]{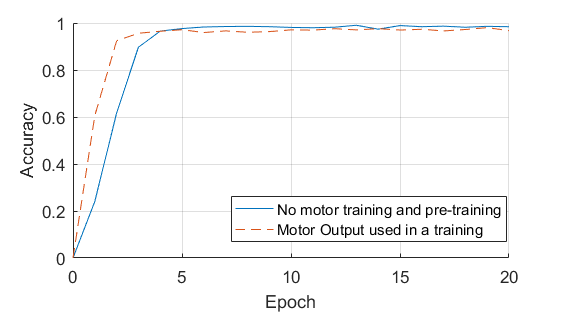}}
	    	\caption{The accuracy development by epoch obtained on the test set for the network trained to reproduce motor positions in parallel with numbers (hyperbolic tangent function used).}
	    	\label{fig:Reconstr_tanh}
	    \end{figure}
        In the next stage of our experiment, where the network was first trained to reproduce motor representation, as can be found in Fig.~\ref{fig:Reconstr_first}, the network was able to learn to recognize the numbers much faster than the network where only the visual part was trained. However, the final accuracy of such training was lower, around 95\% (versus 98\%).
        
        The results where both motor and enumeration outputs were trained in parallel, can be found in Fig.~\ref{fig:Reconstr_tanh}. Such training was beneficial for the network - it can be found that the network learns faster. It is, however, visible that it converges finally to a smaller value compared to the training without motor replication (97\% accuracy versus 98\%) but the difference is not very significant.

    \section{Experiment and analysis on numerosity estimation and subitizing}
    \subsection{Model description}\label{numbers_descr}
    In this section, the same model was used as the one presented in Section \ref{gestures_out}. We provided different input to the model and analyse the output in more details.
    
    \subsection{Training description}\label{train_descr3}
    We used the training method that was described in Section \ref{gestures_out} where the network was pre-trained and the hyperbolic tangent function was used for learning rate for motor training.
    Same as before, the data was composed of batches of size 9, however, the distribution of numbers in these sets was different. There were two types of distributions (similar two types of distributions were used in \cite{chencan}):
    \begin{itemize}
        \item Uniform distribution: The same distribution as used in previous sections. Each of 9 elements in our batch corresponds to a different number from 1 to 9.
        \item Zipfian distribution: It has been found that many types of data in physical and social sciences can be approximated with this type of distribution, it is observed in word frequency distribution \cite{piantadosi2016rational}. In case of number word frequencies, the exponent in the Zipf distribution can be approximated by 2 and so the probability distribution can be express by normalized 1/N$^2$, where N is the value of the number \cite{piantadosi2016rational,chencan}.
    \end{itemize}

     This time, the enumeration output was analysed for each numerosity separately. Thus, in the results, we can see the development of numerosity estimation for each number from 1-9 separately (the accuracy of the recognition of that specific number of objects).
     
     In\cite{chencan}, apart from different architecture, the network was not using sensorimotor data and was not pre-trained in an unsupervised way as it was in our experiment (we also used different visual input - simplified). As we obtained different results, we decided to check our network performance for the condition where no unsupervised pre-training was performed and for the condition with no motor training (not pre-trained and trained to replicate motor data). This allowed to check if these changes in the results are caused by the differences in the training or are only due to a different architecture.
    
    \subsection{Results} \label{numbers_results}
    \begin{figure}[tb]
    	\centering
    	\begin{subfigure}[b]{1\columnwidth}
    		\centering
    		\includegraphics[width = 0.82\columnwidth]{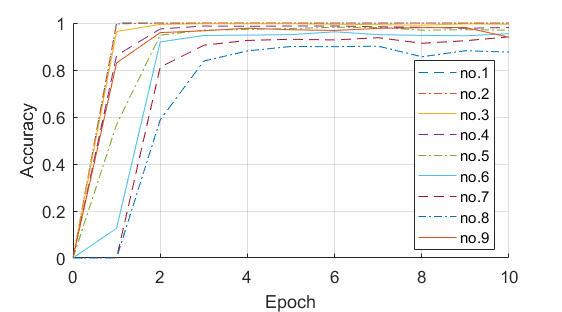}
    		\caption{Uniform distribution of numerosities during training.}
    	\end{subfigure}
    	\begin{subfigure}[b]{1\columnwidth}
    		\centering
    		\includegraphics[width = 0.82\columnwidth]{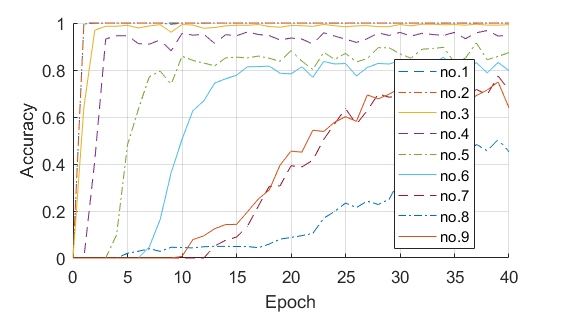}
    		\caption{Zipfian distribution of numerosities during training.}
    	\end{subfigure}
    	\caption{The accuracy development by epoch obtained on the test set for each number of objects presented to the network.}
    	\label{fig:Numbers}
    \end{figure}
    Fig.~\ref{fig:Numbers} shows the development through epochs of accuracy for each number and for two different data distributions (uniform, Zipfian). It can be seen that generally smaller numbers are learned faster than higher ones, even in case of uniform distribution. The only exception from this pattern is number 9 which is the boundary value - the highest number in our training and test sets. As it can be seen the difference in accuracy distribution between the numbers is much more significant when Zipfian distribution is used (this was expected as the exposition of the network to higher numbers is much lower).
    We also found that the network has the tendency to underestimate higher numbers in case of a Zipfian distribution. That was probably also caused by higher exposition to smaller numbers. This is in line with human studies \cite{revkin2008does} and with what was observed in \cite{chencan}.

    The subitizing range was easily observable on the plots with variation coefficient in \cite{revkin2008does}. The variation coefficient is defined as standard deviation of number responses (for each number from 1-9 separately) divided by the mean value of those responses (mean estimated number). According to Weber's law, its value should be constant, with the exception of the subitizing region which has been found to not obey Weber's law. In the subitizing region that value was close to zero \cite{revkin2008does}.
    
    \begin{figure}[tb]
        \centerline{\includegraphics[width = 0.82\columnwidth]{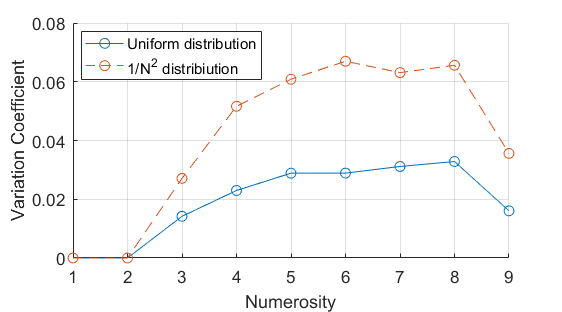}}
        \caption{Variation coefficient as a function of numerosity.}
        \label{fig:Variation_Coef}
    \end{figure}
	\begin{figure}[tb]
		\begin{subfigure}[b]{1\columnwidth}
			\centering
			\includegraphics[width = 0.82\columnwidth]{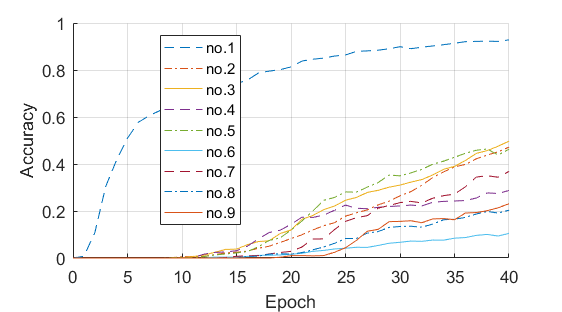}
			\caption{Uniform distribution of numerosities during training.}
		\end{subfigure}
		\begin{subfigure}[b]{1\columnwidth}
			\centering
			\includegraphics[width = 0.82\columnwidth]{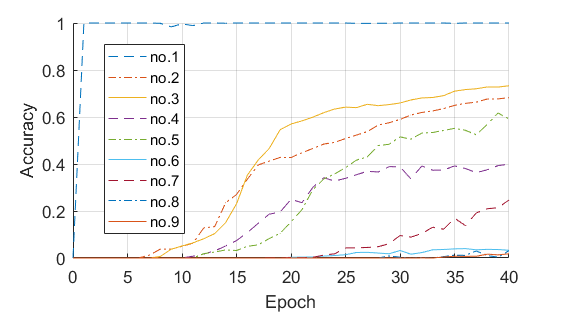}
			\caption{Zipfian distribution of numerosities during training.}
		\end{subfigure}
		\caption{The accuracy development by epoch obtained on the test set for each number of objects presented to the network. Simulations without unsupervised pre-training.}
		\label{fig:Numbers_no_pre}
	\end{figure}
    As can be seen in the Fig.~\ref{fig:Variation_Coef}, in our experiments we observe the region with values of variation coefficient close to zero (for numerosities 1 to 2), also the value of that coefficient is smaller for 3 and 4 compare to number 5 to 8. In the case of number 9, its value is smaller but that might be, again, caused by the fact that 9 is a boundary value\footnotemark.
    \footnotetext{The mean value from 40 simulations obtained on the test set after 15 epochs of training for the uniform distribution and after 40 for Zipfian one. The similar shape was already visible in the earlier stages of the training}
    This smaller value was also observed in human studies \cite{revkin2008does} for the highest test number used in their experiment: 8. This subitizing pattern can be also seen in Fig.~\ref{fig:Numbers} where numbers 1-2 are trained much faster and reach the accuracy close to 100\%. In this figure also number 3 seems to show much higher performance in both speed of convergence and final accuracy. Such subitizing region was not found in \cite{chencan}.

    \begin{figure}[tb]
        \centerline{\includegraphics[width = 0.82\columnwidth]{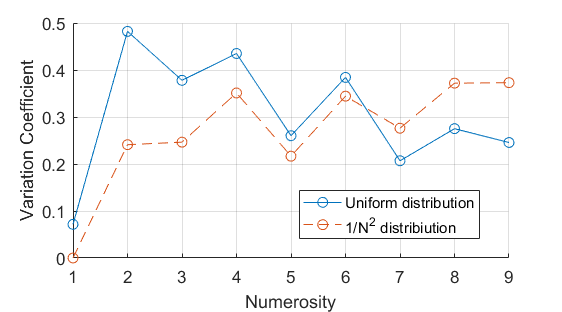}}
        \caption{Variation coefficient as a function of numerosity. Simulations without unsupervised pre-training.}
        \label{fig:Variation_Coef_no_pre}
    \end{figure}

	 \begin{figure}[tb]
		\centerline{\includegraphics[width = 0.82\columnwidth]{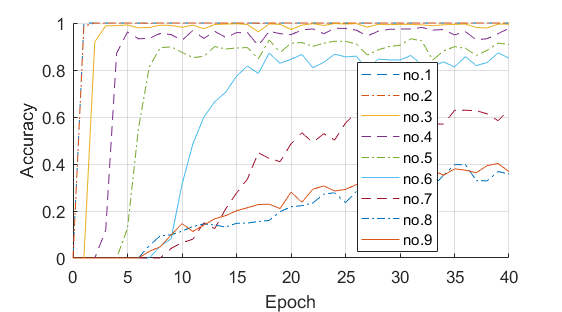}}
		\caption{The accuracy development by epoch obtained on the test set for each numerosity  for Zipfian distribution. No motor output training and pre-training in the simulations.}
		\label{fig:Number_no_motor}
	\end{figure}
    We obtained different results in an experiment where no pre-training was used. In Fig.~\ref{fig:Numbers_no_pre} we can see that such training is significantly better for number 1 (compared to other numbers) in both cases: uniform and Zipfian distributions. The order of development of accuracy of other numerosities is not regular which is different from the results where pre-training was used. If we look at the variation coefficient plot\footnotemark we can not find the subitizing region and only the number 1 is characterised by a smaller value of that coefficient (see Fig.~\ref{fig:Variation_Coef_no_pre}). In this case (lack of pre-training) the obtained results are similar in shape to those in \cite{chencan} where the subitizing region was not visible.
    \footnotetext{The mean value from 40 simulations obtained on the test set after 40 epochs of training}

    The results where sensorimotor data was not used can be seen in Fig.~\ref{fig:Number_no_motor}. We present the plots only for Zipfian distribution as the results for uniform distribution follow the same pattern. In general, it can be seen that the results for small numbers are slower in their accuracy training compare to those where motor data was used (you can see this difference when you compare the curve for number 3 or 4 in Fig.~\ref{fig:Numbers} and \ref{fig:Number_no_motor}). There was no significant difference in the variance coefficient compared to the training with trained motor output (for the plot obtained in the final stage of the training).

    \section*{Acknowledgment}
    We acknowledge the support of EPSRC through project grant EP/P030033/1 (NUMBERS).
    
    \section{Conclusions and discussion}
    In this paper, neuro-robotics models with a deep artificial neural network capable of generating finger positions and number estimation were proposed. The presented studies show an improvement in training speed of number estimation when the network is also producing finger counting in parallel with the estimated number. We can also see that this process of enumeration is strongly boosted when the network is first trained in an unsupervised way layer-by-layer, treating each layer as an autoencoder or RBM. We found autoencoder based pre-training to be a bit more influential in the final training performance. However, both types of unsupervised pre-training methods caused significant improvement.
    
    Moreover, we observed that the network learns much faster to recognise smaller numerosities. Also, the final accuracy for smaller numbers is higher. It was observed that the network learns perfectly to recognise numerosities 1 and 2, which was visible as a zero value region in the variation coefficient plot. This was similar to subitizing observed in humans, where the zero region on the plot is, however, in the range 1-4 \cite{revkin2008does}.
    
    We can hypothesize that when using proper training approach (unsupervised pre-training and use of sensorimotor data) one single model could potentially obtain subitizing in range 1-4 similarly to humans (and some animals) and follow Weber's law for higher numbers. The counterargument could be that the babies not only escape Weber's law in the subitizing range but also the performance for numbers above this limit falls down to a chance level \cite{revkin2008does}. However, this could be also explained in a way that the training is in its early stage and so when looking at the progress of training (Fig.~\ref{fig:Numbers}) we can also observe that higher numbers are practically not trained at all in the first stages of the training.
    
    The results presented in this article provide interesting information on how sensorimotor data can influence the numerosity estimation learning process and how unsupervised pre-training boosts and changes it.
    
    Our model uses simplified visual input, it would be interesting in the future to perform similar tests on more realistic images. Such change might require to add convolutional layers to the network to deal efficiently with the real image. We can, however, assume that even with our simplified visual input the results might follow the same pattern. One of the arguments for that is that the results we obtained, when no unsupervised training was used, are similar in shape to those in \cite{chencan}.

	\bibliographystyle{IEEEtran} 
	\bibliography{ref}

\end{document}